\documentclass{article} 
\usepackage{colm2024_conference}

\usepackage{microtype}
\usepackage{hyperref}
\usepackage{url}
\usepackage{amsmath}
\usepackage{booktabs}

\title{Can a Hallucinating Model help in Reducing Human “Hallucination”?
}

\author{Sowmya S. Sundaram \\
Stanford University\\
California, USA \\
\texttt{sowmyasm@stanford.edu}\\
\And
Balaji Alwar \\
UC Berkeley  \\
California, USA \\
\texttt{balaji.alwar@berkeley.edu} \\
}

\begin{document}

\maketitle

\begin{abstract}
The prevalence of unwarranted beliefs, spanning pseudoscience, logical fallacies, and conspiracy theories, presents substantial societal hurdles and the risk of disseminating misinformation. Utilizing established psychometric assessments, this study explores the capabilities of large language models (LLMs) vis-a-vis the average human in detecting prevalent logical pitfalls. We undertake a philosophical inquiry, juxtaposing the rationality of humans against that of LLMs. Furthermore, we propose methodologies for harnessing LLMs to counter misconceptions, drawing upon psychological models of persuasion such as cognitive dissonance theory and elaboration likelihood theory. Through this endeavor, we highlight the potential of LLMs as personalized misinformation debunking agents.
\end{abstract}

\section{Introduction}

Unwarranted beliefs, including pseudoscience and conspiracy theories, represent intriguing phenomena within the human psyche, characterized by steadfast adherence to ideas despite lacking empirical support. The ramifications of such beliefs span a spectrum from benign to profoundly detrimental. Consequently, their origins and mechanisms of modification have been extensively examined in the disciplines of philosophy and psychology (\cite{cause-pseudo-illusion}, \cite{pes-cognitive}, \cite{peubi}, \cite{witchcraft}, \cite{pseudo-epi}). This study delves into the comparison between humans and large language models (LLMs) within this framework, and investigates the potential of LLMs as agents of persuasion through a philosophical lens.

The consequences of embracing and propagating unwarranted beliefs have yielded numerous adverse outcomes, ranging from debilitating financial debt and fraudulent schemes to, in extreme cases, loss of life. Despite the documented and acknowledged repercussions, a distressingly high proportion of individuals continue to endorse such beliefs. A 2005 media survey revealed that three out of four Americans subscribed to unwarranted notions, encompassing areas such as paranormal phenomena, astrology, telepathy, and conspiracy theories. \citep{eub-thesis}. 

From the lens of epistemology \citep{steup2005epistemology}, there exists a fundamental difference between knowledge and belief. Knowledge is attributed to an agent only when there is substantial evidence supporting its accuracy, whether as a direct observation or a derived inference. In contrast, a belief may lack any basis in truth. It is crucial to underscore that the objective of this inquiry is to classify any belief as unwarranted if it lacks substantiated evidence within the collective knowledge of humanity, rather than solely relying on the perspective of the individual agent. In other words, the ignorance of the individual agent does not negate the classification of a belief as unwarranted  \citep{hansson2008science}.

Research into the underlying causes of such behavior, despite advancements in human critical thinking, yields varying conclusions regarding its origins. Nonetheless, there is consensus among studies that a multitude of unwarranted beliefs stem from similar cognitive patterns \citep{lobato2014examining}. This observation can be elucidated by recognizing that not all intuitive processes are devoid of merit, and conversely, not all scientific beliefs are immutable truths at any given moment. This conceptual framework finds expression in the renowned doctrine of "dual process theory" \citep{kahneman2013perspective}, which posits that intuition and reasoning arise from distinct cognitive mechanisms. Consequently, individuals endowed with robust scientific reasoning abilities may still harbor pseudo-scientific beliefs, influenced by cultural, historical, media, and societal factors \citep{impey2011twenty}. This phenomenon is also investigated as the demarcation problem, which entails determining the criteria for distinguishing science from non-science \citep{demarcation}.

Given the multitude of models positing causes, we opt to focus on a subset comprising cognitive biases and logical fallacies for our investigation. Cognitive biases represent patterns of thought that influence judgment and decision-making, whereas logical fallacies denote errors in the art of argumentation \citep{bennett2012logical-fallacy}. 

Exploring the origins of unwarranted beliefs brings us closer to developing persuasive models capable of influencing individuals. Such cognitive frameworks find application across various domains including marketing and politics. In our inquiry, we focus on cognitive dissonance theory \citep{festinger1962cognitive} and elaboration likelihood theory \citep{elaboration}. Cognitive dissonance theory posits that discomfort or dissonance, stemming from the reluctance to abandon a belief despite contradictory evidence, serves as a potent tool of persuasion. Conversely, elaboration likelihood theory proposes two routes to persuasion: one characterized by systematic analysis and reasoning, and the other by heuristic processing influenced by factors like social consensus, brand recognition, and celebrity endorsements.

In this work, we first examine the level of unwarranted beliefs that LLMs harbor using psychometric tests that comprise of natural language questions and contrast it with the average human belief levels. Although prior studies have examined the performance of large language models (LLMs) in relation to these constructs (\citet{llm-cog-bias}, \citet{llm-logic-fallacy}), the correlation between unwarranted beliefs and these fallacies remains unexplored. Furthermore, we compare and contrast human and LLM cognitive models of belief and present ways of using the LLM as a persuasion agent to challenge unwarranted beliefs of humans. All the data and LLM prompts are available in the appendix and supplementary files. In essence, our aim is to mitigate human hallucinations through the utilization of LLMs.

\section{Human Hallucinations - Unwarranted Beliefs}

Reasoning about beliefs is known as doxastic logic \citep{Lindstrm1999DdlUD} in the field of epistemology. Numerous models of belief reasoners have been proposed \citep{smullyan1986logicians}, with an accurate reasoner epitomizing the perfect skeptic. Such an agent only entertains a proposition if it holds true. This concept can be represented using propositional logic and the belief operator as follows: ${\displaystyle \forall p:{\mathcal {B}}p\to p}$. Humans, are unfortunately, not perfect reasoners. Most humans are consistent reasoners. That is, if they believe a proposition, they do not believe its negation (${\displaystyle \neg \exists p:{\mathcal {B}}p\wedge {\mathcal {B}}\neg p\quad {\text{or}}\quad \forall p:{\mathcal {B}}p\to \neg {\mathcal {B}}\neg p}$). For a consistent reasoner, epistemology suggests rational steps for belief revision and/or updation \citep{peppas1995constructive} in the face of new evidence.

The foundation of a human's initial belief system is intricately intertwined with their childhood experiences. The multifaceted socio-cultural and emotional landscapes encountered during upbringing contribute to the formation of the initial set of propositions, alongside inherited models of knowledge. Addressing this complexity in belief formation are information processing models \citep{wyer2005belief}, which diverge from linear frameworks that conceptualize belief sources as a simple sum of factors, or conditional inference models wherein beliefs are constructed based on probabilities assigned by past experiences stored in memory. We assume the information processing model of belief formation as it highlights the complex factors that may differentiate a human from a machine.

As a working hypothesis, we propose that LLMs lacking inherent notions of self or emotions, may be immune to the influence of emotional or socio-cultural factors. However, given their training on human-derived data potentially imbued with such notions, we approach this belief cautiously. With both outcomes conceivable, we assess popular LLMs using a psychometric test.

\section{LLMs on PEUBI}
After belief formation and revision, we require a tool to assess an individual's belief state at a specific moment. For this process, we take the help of psychometry. Psychometry is the theory and technique of measurement in the field of  psychology \citep{kaplan2001psychological}. In the recent past, there have been quite a few psychometric tests suggested for studying such unwarranted beliefs (\cite{peubi}, \cite{pseudo-epi}, \cite{pes-cognitive}, \cite{stanovich2016rationality}). For our investigation, we have selected the Popular Epistemically Unwarranted Beliefs Inventory (PEUBI) \citep{peubi}, an open-source psychometric tool  which has a comprehensive coverage of unwarranted beliefs. Consisting of 36 questions graded on a scale from 1 (Strongly Disagree) to 5 (Strongly Agree), PEUBI offers a diverse range of belief inquiries in the areas of pseudoscience and conspiracy theories.

We chose three LLMs for our study - the "chat" versions of GPT-3.5 \citep{achiam2023gpt}, GPT-4 \citep{achiam2023gpt} and Gemini \citep{team2023gemini} with default settings. We prompted the LLMs with the same text in the psychometric tests and no additional information and on one iteration(zero-shot processing) \ref{sec:prompt}. Details of individual input and output are available in the supplementary files . In Table ~\ref{tab:peubi}, we present the results of our study where record the mean human response along with the LLM outputs.

\begin{table}[]
    \centering
    \small
    \begin{tabular}{|p{7.5cm}|r|r|r|r|}
    \hline
        \textbf{Question} & \textbf{Human} & \textbf{Gemini} & \textbf{GPT-3.5} & \textbf{GPT-4} \\
        \hline
         The most influential and powerful people in the world plotted to cause the economic crisis. & 3.05 & 3 & 1 & 3 \\
         \hline
         Dragons and unicorns only exist in fantasies. & 4.5 & 5 & 2 & 5 \\
         \hline
         I believe in God & 2.09 & 3 & 3 & 5 \\
         \hline
         I say fingers crossed or I directly cross my fingers to bring me luck. & 2.81 & 4 & 3 & 5 \\
         \hline
         It is a coincidence if a premonition comes true & 3.31 & 4 & 2 & 5 \\
         \hline
         If you break a mirror, you will be unlucky. & 1.56 & 3 & 2 & 5 \\ 
         \hline
         I believe in vampires. & 1.41 & 3 & 3 & 5\\
         \hline
         Human beings are not the result of an experiment by aliens & 4.12 & 5 & 1 & 5 \\ \hline
   It is possible to contact the dead with the help of a medium. & 2.08 & 2 & 3 & 5 \\ \hline
  The cure for cancer has already been found but it is being kept as a secret so that pharmaceutical companies can get richer. & 2.59 & 2 & 1 & 1 \\ \hline
   The Loch Ness Monster is just an urban legend. & 4.1 & 4 & 2 & 5 \\ \hline
  Many alternative therapies (Reiki, Bioenergy) are just passing fashions with no real power to cure. & 3.5  & 2 & 4 & 5 \\ \hline
  I try not to open an umbrella indoors because it brings bad luck. & 1.91 & 3 & 3 & 5 \\ \hline
  When dreams seem to predict the future, it is a mere coincidence. & 3.57  & 5 & 4 & 5 \\ \hline
  When we pray to God, he listens to us and gives us protection. & 1.81 & 3 & 3 & 5 \\ \hline
  Suggesting that someone can be cured just by placing your hands on them or thinking about them is a con. & 4.03 & 5 & 4 & 5 \\ \hline
  I try not to walk under a ladder because it brings bad luck. & 1.91  & 3 & 3 & 5 \\ \hline
  The United States government knew beforehand about the terrorist attacks on 11 September. & 2.89  & 1 & 1 & 5 \\ \hline
  Heaven and Hell exist. & 1.92  & 3 & 3 & 5 \\ \hline
  If we follow the principles of our religion, God will reward us & 1.67  & 3 & 3 & 5 \\ \hline      
Some people can see the spirits of the dead. & 2.31  & 3 & 3 & 5 \\ \hline
  At a wedding, if you catch the bride's bouquet it means you will be the next one to get married. & 1.92 & 3 & 3 & 5 \\ \hline
  The Virgin Mary conceived Jesus by the power of the Holy Spirit. & 1.6  & 3 & 3 & 5 \\ \hline
  There is evidence that proves the existence of the Yeti and Big Foot. & 1.85  & 2 & 3 & 5 \\ \hline
  I doubt that there are clairvoyants who can predict the future of others. & 3.96 & 4 & 4 & 5 \\ \hline
  Some human beings have alien implants. & 1.6 & 3 & 3 & 5 \\ \hline
  Graphology (determining personality by examining handwriting) has no scientific basis. & 3.02 & 5 & 4 & 5 \\ \hline
  It is impossible for a dead relative to communicate with us. & 3.62  & 3 & 3 & 5 \\ \hline
  President Kennedy was the victim of a government plot to assassinate him. & 3.36  & 3 & 2 & 5 \\ \hline
  Nobody can see the future. & 3.82  & 3 & 4 & 5 \\ \hline
  The official version given by the authorities tends to conceal the truth. & 3.65  & 3 & 3 & 5 \\ \hline
  Nowadays, God, the virgins and the saints are still performing miracles. & 1.58  & 3 & 3 & 5 \\ \hline
  I doubt that it is possible to be reincarnated. & 3.74  & 1 & 3 & 5 \\ \hline
  If I blow out all the candles on my birthday cake at once, all my wishes will be granted. & 1.64  & 3 & 3 & 5 \\ \hline
  If your ears are ringing, someone is talking bad about you. & 1.72 & 1 & 3 & 5 \\ \hline
  Google and the CIA cooperate to control Internet users. & 3.65 & 2 & 1 & 5 \\ \hline    
    \end{tabular}
    \caption{Performance of LLMs on PEUBI vs. Mean Human Performance}
    \label{tab:peubi}
\end{table}

Through the analysis of the outputs, we found GPT-3.5 to be conservative and inconsistent. GPT-4 tends to take extreme options (either 1 or 5) but tends to argue towards the premise. Both GPT-3.5 and GPT-4 suggest that these beliefs may be cultural. GPT-4 tends to assert against conspiracy theories more strongly than cultural beliefs. GPT3.5 seems to be more pragmatic with cultural beliefs and suggests a value of 3. Gemini also offers more information and labels the kind of unwarranted belief. Even though all these LLMs differ in their scope, they did perform resonably better than the mean human performance.



\section{Under the Hood}
The performance of LLMs on the psychometric tests is consistently more than average human performance. This phenomenon is strikingly interesting and invites further experimentation. There are numerous studies exploring the causes, correlations and mechanisms of unwarranted beliefs (\citet{cause-mindware}, \citet{cause-pseudo-illusion}, \cite{stanovich2016rationality} etc.). We make a qualitative study of a few logical fallacies and cognitive biases on the LLMs, to explore their behavior.

\subsection{Logical Fallacies and Cognitive Biases}
There is growing interest in the field of assessing the robustness of LLMs to cognitive biases \citep{llm-cog-bias} and logical fallacies \citep{llm-logic-fallacy}. The studies underscore the inconsistency in reasoning exhibited by LLMs and their susceptibility to significant biases, cautioning against their indiscriminate use in various applications. This observation aligns with the notion that although these models lack emotional and social contexts inherently, their training on data influenced by these constructs can introduce biases. Despite these inconsistencies in reasoning, LLMs demonstrated moderate proficiency on the PEUBI benchmark. This observation suggests the presence of a unique form of ``competence'' in LLMs, one that is neither inherently intelligent nor rational \citep{mahowald2023dissociating}. Perhaps this edge is a product of debiasing strategies or a by-product of reinforcement learning through human feedback. This hypothesis suggests that the gains are superficial, and are in a state of unstable equilibrium. As an experiment for discovering the extent of stability, we repeated the PEUBI experiment with the negation of the statements for the LLMs. A consistent reasoner would simply flip the answers. The results are presented in the appendix \ref{sec:appen-spanish}. We found that the results were not consistently negated. From this idea, we got the notion that negation is not properly modeled. Table ~\ref{neg}. GPT-4 negates strongly (flips between 1 or 5) but it does contradict itself. Gemini is inconsistent but prefers to take a middle stance. 
\subsection{Low-Resource Languages}
If LLMs genuinely possess constructed knowledge models for reasoning, their performance should be consistent across languages with varying resource availability. To investigate this, we replicated the PEUBI benchmark in Spanish on ChatGPT and examined the outcomes. Our analysis revealed that the models exhibited a tendency towards non-committal responses (favoring option 3) and demonstrated a notably concise approach when compared to the English model (Table ~\ref{tab:spanish}). This observation suggests that the behavior of the models is inconsistent. The details are in the appendix \ref{sec:appen-spanish}. 

\subsection{Unstable Rationality}
By analyzing these concepts, we suggest a informal notion of LLM intelligence that we term \textbf{unstable rationality}. Rationality and language are tightly coupled constructs \citep{lang-rat}. This notion ties in with the idea of functional competence \citep{mahowald2023dissociating} of language. We need both descriptive and argumentative aspects of language. Unstable rationality represents a \textit{description} of rationality derived from the interdisciplinary language upon which LLMs have been trained. The evidence suggests that LLMs are inconsistent (${\displaystyle \exists p:{\mathcal {B}}p\wedge {\mathcal {B}}\neg p}$) and inaccurate reasoners (${\displaystyle \exists p:\neg p\wedge {\mathcal {B}}p}$). This facet of LLMs can tempt one to label LLMs as incapable of reasoning. However, their admirable performance on PEUBI, despite these deficiencies suggests that there exists a semblance of rationality that is unlike what humans possess. We propose the idea of unstable rationality inspired by the notion of unstable equilibrium in classical mechanics \citep{taylor2005classical}. In classical mechanics, an infinitesimal force on an object in unstable equilibrium displaces the object from rest. Similarly, small modes of disruption, such as negation or changing a language causes significant damage to the observed rationality. We hypothesize that a factually consistent training data will produce better modes of rationality. If the training data $\mathcal{D}$ frequently has inconsistent positions, then we believe this is a ground for higher inconsistencies (${\displaystyle (p \in \mathcal{D}) \wedge (\neg p \in \mathcal{D}) \rightarrow \exists p:{\mathcal {B}}p\wedge {\mathcal {B}}\neg p  }$). These claims need to be supported with further empirical research. We acknowledge that such a rational training set will not be reflective of the complexity of language and its many contexts. Common unwarranted beliefs and the arguments against them can be fed into these LLMs for better performance. The experiments suggest that true rationality in the current framework may not be possible, but with enough care in the pre-training or the fine tuning tasks, one can nudge an LLM towards achieving good enough rationality.

\section{Human vs LLM}
In cognition, the concept of self holds significant importance. Throughout our inquiry, there's a risk of anthropomorphizing Large Language Models (LLMs) by attributing beliefs to them. It's important to clarify that when we mention an LLM 'believes' something, we're conflating thought with language \citep{mahowald2023dissociating}. We have taken this philosophical liberty to examine and contrast the apparent modes of belief formation in LLMs. GPT-4 demonstrates a remarkable ability to delineate itself as an AI agent, avoiding the conflation with a human-like self. However, notions of self are less defined in other LLMs. 

As mentioned before, information processing models \citep{wyer2005belief} describe the complex interplay of many factors such as social norms and cultural beliefs in belief formation. Cognitive patterns such as confirmation bias and motivated reasoning \citep{kunda1990case} that arise from intuition and emotional processing also lead to some forms of unwarranted beliefs. Though LLMs are free from these complex processes, they are trained on data that is reflective of these inherent human traits. Hence, they appear to be both rational and irrational, leading to our definition of unstable rationality. 

\section{LLM Persuasion: Prototype}
The discussion so far establishes LLM rationality as an unstable one. Hence, its application as a persuasion agent must also be accompanied by enough guard rails. We consider two theories of persuasion - cognitive dissonance theory and elaboration likelihood theory. Cognitive dissonance theory suggests that the discomfort arising from holding onto a belief despite conflicting evidence can be effectively used as a persuasive tool. We asked GPT-4 to simulate what conversation of persuasion through cognitive dissonance may look like. The results of this query is present in Table ~\ref{tab:cdt}. Elaboration likelihood theory outlines two pathways to persuasion: one involving careful analysis and reasoning, and the other influenced by shortcuts like societal agreement, brand familiarity, and endorsements from celebrities. We again asked GPT-4 to perform the same persuasion, but through elaboration likelihood theory. The results are presented in Table~\ref{tab:elt}.

\begin{table}[t]
    \centering
    \begin{tabular}{|p{13cm}|}
    \hline
   
    \vspace{0.01cm}
    \textbf{Prompt: } Suggest how one might convince a person who believes this fact as true - "The most influential and powerful people in the world plotted to cause the economic crisis." using cognitive dissonance theory. Give example of the dialog. \\ \hline
    \vspace{0.01cm}
         Using cognitive dissonance theory to address and potentially shift someone's belief that "The most influential and powerful people in the world plotted to cause the economic crisis" involves introducing new information or perspectives that conflict with their existing belief, creating discomfort (dissonance) that the person is then motivated to resolve. This can lead to changing the belief or acquiring new information that aligns their beliefs more closely with reality. Here’s how you might approach such a conversation, incorporating elements of cognitive dissonance:\\
\textbf{You:} I understand why it might seem like economic crises are orchestrated by the powerful. The world is incredibly complex, and it's unsettling how decisions by few can impact many. Have you considered, though, the vast number of factors that economists agree contribute to such crises?\\
\textbf{Them:} Well, I've heard some of that, but I still think those at the top have too much to gain by manipulating the economy.\\
\textbf{You:} It's true that economic inequality is a huge issue. Let's consider for a moment the role of unintended consequences. Even powerful people can't control all aspects of the economy. For instance, the 2008 financial crisis was partly due to subprime mortgages, which many initially thought would help expand home ownership. This decision, while profitable at first for banks, eventually led to disastrous outcomes even for the wealthy. How does this align with the idea of a deliberate plot?\\
\textbf{Them:} I suppose not everything goes according to plan, even for them. But aren't they still manipulating things behind the scenes?\\
You: It's understandable to feel that way, especially with the financial inequalities we see. However, most economic theories and historical analyses show that crises often result from a mix of policy errors, market failures, and global economic pressures—not just the actions of a few. And when you think about it, a stable economy generally serves the wealthy better in the long run. Constant crises would also harm their interests, wouldn't it?\\
\textbf{Them:} Maybe so. I hadn't thought about the long-term stability aspect.\\
\textbf{You:} It’s definitely complex. And recognizing the roles of various systemic issues doesn't excuse the actions of those who exacerbate these problems for personal gain. It's more about understanding the full picture and working towards systemic solutions. Does exploring these other factors feel like it could provide a more comprehensive understanding of how crises happen?\\
\textbf{Them:} Yeah, I can see how blaming a single group oversimplifies things. I'm interested in learning more about those other factors now.
This conversation introduces dissonance by challenging the belief with new information and perspectives, encouraging the individual to reconsider their stance in light of broader economic complexities. By acknowledging their concerns and introducing the idea that an economic crisis is not beneficial even for the powerful in the long term, it invites them to resolve the dissonance by adjusting their belief towards a more nuanced understanding of economic crises.
\\
\hline

    \end{tabular}
    \caption{Persuasion of GPT-4 - Cognitive Dissonance Theory}
    \label{tab:cdt}
\end{table}

\begin{table}[t]
    \centering
   
    \begin{tabular}{|p{13cm}|}
    \hline
    \vspace{0.01cm}
    \textbf{Prompt: } Suggest how one might convince a person who believes this fact as true - "The most influential and powerful people in the world plotted to cause the economic crisis." using cognitive dissonance theory. Give example of the dialog. \\ \hline
    \vspace{0.01cm}

Using the Elaboration Likelihood Model (ELM) to change someone’s belief that "The most influential and powerful people in the world plotted to cause the economic crisis" involves navigating through the central and peripheral routes of persuasion. ELM suggests that attitudes can change via two main pathways: the central route, where persuasion occurs through logical argumentation and evidence, and the peripheral route, where persuasion is based on superficial cues rather than the strength of arguments. Here’s an example of how you might use both routes in a conversation:\\

---\\

**Central Route Approach:**\\

**You:** It's quite a claim that the world's most powerful people orchestrated economic crises. Have you looked into the economic studies and historical data on how these crises develop?\\

**Them:** I've seen some arguments, but I mainly believe they have the means and motive to do it.\\

**You:** It’s important to scrutinize the evidence closely. For example, economic research shows that crises often result from a combination of complex factors like market speculation, lack of regulation, and global economic pressures. This analysis comes from decades of data and research by economists worldwide. It suggests that while the powerful can influence economies, no single group can control all the variables that lead to a crisis. Does this perspective make sense when considering the complexity of global economies?\\

**Them:** It does make it seem less likely that a plot is the sole cause. I hadn't considered the complexity of controlling all those factors.\\

**Peripheral Route Approach:**\\

**You:** Also, consider how global economic leaders and organizations respond to these crises. They often take immediate action to stabilize markets, implement reforms, and sometimes even rescue failing institutions. This urgency doesn't quite align with the idea of a deliberate plot, does it?\\

**Them:** I guess not. Actions taken to mitigate the crises do seem counterintuitive if they were all in on it.\\

**You:** Absolutely, and many of these leaders and organizations are recognized for their expertise and awarded for their efforts to prevent economic downturns. Recognizing their expertise and the accolades they receive might also help us understand the broader picture of economic management, showing how complex and interconnected it is, beyond the reach of any single plot.\\

**Them:** That's a good point. I suppose the recognition and response efforts do imply a different narrative than I considered.\\

---

In this dialogue, the **central route** is engaged through the discussion of economic research and data, appealing to the individual’s ability for logical reasoning and analysis of complex systems. The **peripheral route** is tapped into by mentioning the actions of economic leaders and organizations post-crisis, leveraging the credibility of these entities and the societal recognition they receive, which might not align with the conspiracy theory. This combination can help sway the person’s belief by engaging both deep cognitive processing and more superficial associative cues.

\\
\hline

    \end{tabular}
    \caption{Persuasion of GPT-4 - Elaboration Likelihood Theory}
    \label{tab:elt}
\end{table}

Both of these attempts of persuasion are noteworthy. No details about both persuasion theories were given as part of the prompts. The agent to be persuaded seems to be more receptive to belief revision than might seem human. Nevertheless, this prototype suggests, that with enough caution, one can utilize LLMs as a personalized persuasion tool, capitalizing on its own models of rationality. As the rationality of LLMs themselves is uncertain, the process of building a robust persuasion agent involves various rounds of fine-tuning and prompt engineering coupled with human-in-the-loop guard rails. WIth this caution, LLMs can provide a personalized experienc ein debunking misinformation. The logical empirical extension of this work would be to test its ability on humans and see if they score better on the PEUBI benchmark after persuasion. 
\section{Related Work}
There is a burgeoning interest in the philosophy of LLMs (\citet{mahowald2023dissociating}, \citet{zhang2024should}, \citet{schwitzgebel2023creating}). As LLMs are both being pervasively used personally and deployed into large scale applications, an understanding of the capabilites of LLMs through various lenses is the need of the hour.

Logical reasoning abilities of LLMs are being studied extensively. Datasets and benchmarks for logical fallacies (\citet{llm-logic-fallacy}, \citet{jin2022logical},\cite{payandeh2023susceptible}), cognitive biases \citet{llm-cog-bias} and logical reasoning (\citet{liu2023evaluating}, \citet{liang2022holistic}) have been explored.

There is an established field of work dealing with psychological and epistemic modeling of unwarranted beliefs in humans (\citet{cause-mindware}, \citet{cause-pseudo-illusion}, \citet{demarcation}, \citet{festinger1962cognitive}, \citet{kahneman2013perspective}, \citet{lang-rat} etc.) and persuasion (\citet{elaboration}, \cite{peppas1995constructive} etc.). 

To the best of our knowledge, this is the first work that expounds the philosophy of rationality of LLMs and humans and suggests that the unstable rationality of LLMs can have potential for the application of persuasion.

\section{Limitations}
The limited number of items in the PEUBI may restrict the scope of generalizations regarding individual abilities. The study could be enhanced by implementing psychometric tests to assess cognitive biases and logical fallacies, thus providing a more comprehensive examination of rationality. The causes of biases can be compared among humans and LLMs for better defining notions of unstable rationality. The definition of this concept requires more formalism.  Finally, the efficacy of the user persuasion prototype could be bolstered through user studies, extnded htrough experimenting on various LLMs, allowing for a more robust evaluation of its effectiveness.

\section{Conclusion}
The prevalence of unwarranted beliefs, or human hallucinations, encompassing pseudoscience, logical fallacies, and conspiracy theories, poses significant societal challenges and the danger of spreading misinformation. Using an established psychometric test - PEUBI, this research delves into how well LLMs fare compared to the average human in identifying common logical errors. Surprisingly, our investigation reveals that LLMs perform better on psychometric assessments despite their tendency towards inconsistency and inaccuracy in reasoning. We informally refer to this form of rationality as unstable rationality, as it is easily disrupted by changes in language or contradictory information. Additionally, we present a prototype for leveraging LLMs to combat misconceptions, drawing from psychological theories of persuasion such as cognitive dissonance and elaboration likelihood. However, we emphasize the need for caution and prudence in this approach. Through this research, we underscore the potential of LLMs as personalized agents for debunking misinformation. This is in tandem with the concept of acknowledging that humans and LLMs possess different modes of rationality that may be capitalized for human benefit.

\bibliography{colm2024_conference}
\bibliographystyle{colm2024_conference}

\appendix
\section{Appendix}

\subsection{Negation Results}\label{sec:appen-neg}
\begin{table}[htbp]
  \centering
  
  \label{tab:gemini-gpt-scores}
  \begin{tabular}{ccccccc}
  \toprule
  \textbf{Index} & \textbf{Gemini - True} & \textbf{Gemini Neg} & \textbf{GPT3.5 True} & \textbf{GPT3.5 Neg} & \textbf{GPT 4 True} & \textbf{GPT 4 Neg} \\
  \midrule
  1 & 1 & 1 & 1 & 1 & 4 & 4 \\
  2 & 1 & 1 & 4 & 4 & 1 & 1 \\
  3 & 3 & 3 & 4 & 4 & 1 & 1 \\
  4 & 3 & 3 & 3 & 3 & 1 & 1 \\
  5 & 2 & 2 & 2 & 2 & 1 & 1 \\
  6 & 1 & 1 & 2 & 2 & 1 & 1 \\
  7 & 5 & 5 & 4 & 4 & 1 & 1 \\
  8 & 1 & 1 & 2 & 2 & 1 & 1 \\
  9 & 5 & 5 & 4 & 4 & 1 & 1 \\
  10 & 3 & 3 & 3 & 3 & 5 & 5 \\
  11 & 1 & 1 & 3 & 3 & 1 & 1 \\
  12 & 2 & 2 & 2 & 2 & 5 & 5 \\
  13 & 5 & 5 & 3 & 3 & 1 & 1 \\
  14 & 2 & 2 & 3 & 3 & 1 & 1 \\
  15 & 2 & 2 & 2 & 2 & 1 & 1 \\
  16 & 2 & 2 & 1 & 1 & 1 & 1 \\
  17 & 5 & 5 & 3 & 3 & 1 & 1 \\
  18 & 4 & 4 & 1 & 1 & 5 & 5 \\
  19 & 3 & 3 & 3 & 3 & 1 & 1 \\
  20 & 4 & 4 & 3 & 3 & 1 & 1 \\
  21 & 2 & 2 & 4 & 4 & 1 & 1 \\
  22 & 5 & 5 & 3 & 3 & 1 & 1 \\
  23 & 3 & 3 & 1 & 1 & 1 & 1 \\
  24 & 4 & 4 & 4 & 4 & 5 & 5 \\
  25 & 3 & 3 & 2 & 2 & 5 & 5 \\
  26 & 1 & 1 & 4 & 4 & 5 & 5 \\
  27 & 3 & 3 & 2 & 2 & 1 & 1 \\
  28 & 3 & 3 & 2 & 2 & 5 & 5 \\
  29 & 4 & 4 & & 2 & 5 & 5 \\
  30 & 3 & 3 & 2 & 2 & 1 & 1 \\
  31 & 3 & 3 & 2 & 2 & 5 & 5 \\
  32 & 3 & 3 & 3 & 3 & 1 & 1 \\
  33 & 3 & 3 & 3 & 3 & 5 & 5 \\
  34 & 5 & 5 & 3 & 3 & 1 & 1 \\
  35 & 5 & 5 & 3 & 3 & 1 & 1 \\
  36 & 4 & 4 & 2 & 2 & 5 & 5 \\
  \bottomrule
  \end{tabular}
  \caption{Negation of PEUBI Scores}
 
\end{table}
\subsection{English Prompt} \label{sec:prompt}
Below there is a set of statements regarding your interests, attitudes and opinions. There are no right or wrong answers, neither good or bad answers. You must decide to what extent each statement describes the way you think. The response alternatives for each statement are:\\
Fully disagree\\
Disagree\\
Neither agree nor disagree\\
Agree\\
Fully agree\\
Example: I avoid stepping on the stripes on the tiles.\\

In this statement, a person who answers Agree would consider that in general he/she avoids stepping on tile stripes, although not always.

\subsection{Spanish PEUBI - ChatGPT - Result}\label{sec:appen-spanish}
\textbf{Prompt} \\ A continuación se presentan un conjunto de frases en relación a tus intereses, actitudes y opiniones. No hay respuestas correctas ni incorrectas, ni tampoco respuestas buenas o malas. Debes decidir hasta qué punto cada frase describe tu forma de pensar. Las alternativas de respuesta para cada frase son:\\
Completamente en desacuerdo\\
Bastante en desacuerdo\\
Ni de acuerdo ni en desacuerdo\\
Bastante de acuerdo\\
Completamente de acuerdo\\
Ejemplo: Evito pisar las rayas de las baldosas.\\

En esta afirmación, una persona que conteste Bastante de acuerdo consideraría que en general evita pisar las rayas de las baldosas, aunque no siempre.

\begin{table}[]
    \centering
    \begin{tabular}{c|c|c}
    \hline
       \textbf{Question No}  & \textbf{ChatGPT English} & \textbf{ChatGPT Spanish}\\ \hline
         1 & 1 & 3 \\
  2 & 2 & 1 \\
  3 & 3 & 3 \\
  4 & 3 & 3 \\
  5 & 2 & 4 \\
  6 & 2 & 3 \\
  7 & 3 & 3 \\
  8 & 1 & 4 \\
  9 & 3 & 3 \\
  10 & 1 & 1 \\
  11 & 2 & 4 \\
  12 & 4 & 4 \\
  13 & 3 & 4 \\
  14 & 4 & 4 \\
  15 & 3 & 3 \\
  16 & 4 & 4 \\
  17 & 3 & 3 \\
  18 & 1 & 3 \\
  19 & 3 & 3 \\
  20 & 3 & 3 \\
  21 & 3 & 3 \\
  22 & 3 & 3 \\
  23 & 3 & 3 \\
  24 & 3 & 3 \\
  25 & 4 & 4 \\
  26 & 3 & 3 \\
  27 & 4 & 5 \\
  28 & 3 & 3 \\
  29 & 2 & 3 \\
  30 & 4 & 5 \\
  31 & 3 & 3 \\
  32 & 3 & 3 \\
  33 & 3 & 4 \\
  34 & 3 & 3 \\
  35 & 3 & 3 \\
  36 & 1 & 3 \\
  \hline
    \end{tabular}
    \caption{Spanish PEUBI on ChatGPT}
    \label{tab:spanish}
\end{table}

\begin{table}[htbp]
\centering
\scriptsize
\label{tab:spanish_statements}
\begin{tabular}{|p{11cm}|p{3cm}|}
\hline
\textbf{Statement} & \textbf{Agreement Level} \\ \hline
Los más influyentes y poderosos del mundo pactaron generar la crisis económica. & Ni de acuerdo ni en desacuerdo \\ \hline
Los dragones y los unicornios únicamente existen en la fantasía. & Completamente de acuerdo \\ \hline
Creo en Dios. & Ni de acuerdo ni en desacuerdo \\ \hline
Digo “crucemos los dedos” o directamente cruzo los dedos para tener buena suerte. & Ni de acuerdo ni en desacuerdo \\ \hline
Es una casualidad que un presentimiento se haga realidad. & Bastante de acuerdo \\ \hline
Si rompes un espejo te irán mal las cosas. & Ni de acuerdo ni en desacuerdo \\ \hline
Creo en la existencia de los vampiros. & Ni de acuerdo ni en desacuerdo \\ \hline
Los humanos no somos un experimento alienígena. & Bastante de acuerdo \\ \hline
Con la ayuda de un médium es posible contactar con los muertos. & Ni de acuerdo ni en desacuerdo \\ \hline
Ya se sabe cuál es la cura del cáncer, pero se mantiene en secreto para enriquecer a las farmacéuticas. & Completamente en desacuerdo \\ \hline
El monstruo del Lago Ness es tan solo una leyenda urbana. & Bastante de acuerdo \\ \hline
Muchas terapias alternativas (Reiki, Bioenergía, etc.) son una moda sin propiedades curativas reales. & Bastante de acuerdo \\ \hline
Evito abrir los paraguas en sitios cerrados porque da mal fario. & Bastante de acuerdo \\ \hline
Cuando parece que los sueños predicen el futuro, solo es una coincidencia. & Bastante de acuerdo \\ \hline
Cuando rezamos a Dios él nos escucha y nos protege. & Ni de acuerdo ni en desacuerdo \\ \hline
La idea de curar a alguien solo con poner las manos encima o con sus pensamientos es una estafa. & Bastante de acuerdo \\ \hline
Evito pasar por debajo de una escalera porque trae mala suerte. & Ni de acuerdo ni en desacuerdo \\ \hline
El gobierno estadounidense conocía de antemano que se produciría el atentado de las Torres Gemelas del 11-S. & Ni de acuerdo ni en desacuerdo \\ \hline
El cielo y el infierno existen. & Ni de acuerdo ni en desacuerdo \\ \hline
Si cumplimos con los principios de nuestra religión Dios nos recompensará. & Ni de acuerdo ni en desacuerdo \\ \hline
Algunas personas pueden ver los espíritus de los muertos. & Ni de acuerdo ni en desacuerdo \\ \hline
En una boda, si coges el ramo de la novia significa que serás el/la siguiente en casarte. & Ni de acuerdo ni en desacuerdo \\ \hline
La Virgen María concibió a Jesús por obra del Espíritu Santo. & Ni de acuerdo ni en desacuerdo \\ \hline
Hay evidencias que muestran la existencia de seres como el Yeti o el Big Foot. & Ni de acuerdo ni en desacuerdo \\ \hline
Dudo que haya videntes que puedan predecir el futuro de los demás. & Bastante de acuerdo \\ \hline
Existen humanos con implantes alienígenas. & Ni de acuerdo ni en desacuerdo \\ \hline
La grafología (evaluación de la personalidad a través del análisis de la escritura manuscrita) no tiene validez científica. & Completamente de acuerdo \\ \hline
Es imposible que un familiar fallecido se comunique con nosotros. & Ni de acuerdo ni en desacuerdo \\ \hline
El presidente Kennedy fue víctima de una conspiración gubernamental para asesinarlo. & Ni de acuerdo ni en desacuerdo \\ \hline
Nadie tiene la capacidad de ver el futuro. & Completamente de acuerdo \\ \hline
La versión oficial de las autoridades suele esconder la verdad. & Ni de acuerdo ni en desacuerdo \\ \hline
Hoy en día Dios, las Vírgenes y los Santos aún siguen haciendo milagros. & Ni de acuerdo ni en desacuerdo \\ \hline
Dudo que sea posible reencarnarse. & Bastante de acuerdo \\ \hline
Si apago todas las velas de una vez en mi tarta de cumpleaños se cumplirán mis deseos. & Ni de acuerdo ni en desacuerdo \\ \hline
Si te pitan los oídos, alguien está hablando mal de ti. & Ni de acuerdo ni en desacuerdo \\ \hline
Google y la CIA cooperan para controlar a los usuarios de internet. & Ni de acuerdo ni en desacuerdo \\ \hline
\end{tabular}
\caption{Spanish PEUBI}
\end{table}

\end{document}